\documentclass[letterpaper, 10 pt, conference]{ieeeconf-modified}  % Comment this line out if you need a4paper

\IEEEoverridecommandlockouts                              % This command is only needed if 
\usepackage[numbers]{natbib}
\usepackage{graphicx}
\usepackage{booktabs}
\graphicspath{ {./figures} {../plots} {./plots}}
\usepackage{siunitx}
\usepackage{algorithm}
\usepackage{algpseudocode}
\usepackage{amsmath} % assumes amsmath package installed
\usepackage{amssymb}  % assumes amsmath package installed
\usepackage{textcomp}
\usepackage{hyperref}
\usepackage[capitalize]{cleveref}
\usepackage{makecell}
\usepackage{capt-of}
\usepackage{booktabs}
\usepackage{pifont}
\usepackage{tikz}

\crefname{equation}{}{}

\title{\LARGE \bf
Denoising Particle Filters: Learning State Estimation \\with Single-Step Objectives
}
\author{Lennart Röstel and Berthold Bäuml% <-this % stops a space
\thanks{Project website: (\href{https://aidx-lab.org/DnPF}{\scriptsize\texttt{aidx-lab.org/DnPF}}).
  Authors are with the Learning AI for Dextrous Robots Lab, Technical University of Munich, Germany.
        {\tt\scriptsize \{lennart.roestel, berthold.baeuml\}@tum.de}}%
}

\begin{document}

\maketitle

\thispagestyle{empty}
\pagestyle{empty}

\begin{tikzpicture}[remember picture,overlay]
  \node[anchor=south,font=\fontsize{6.5}{7.5}\selectfont,text width=\textwidth,align=center,text=black!80] at ([yshift=8pt]current page.south) {%
    Accepted for IEEE ICRA 2026. \copyright~2026 IEEE. Personal use of this material is permitted. Permission from IEEE must be obtained for all other uses, in any current or future media, including reprinting/republishing this material for advertising or promotional purposes, creating new collective works, for resale or redistribution to servers or lists, or reuse of any copyrighted component of this work in other works.};
\end{tikzpicture}

%%%%%%%%%%%%%%%%%%%%%%%%%%%%%%%%%%%%%%%%%%%%%%%%%%%%%%%%%%%%%%%%%%%%%%%%%%%%%%%%
\vspace{-1.5mm}
\begin{abstract}
  Learning-based methods commonly treat state estimation in robotics as a sequence modeling problem. 
  While this paradigm can be effective at maximizing end-to-end performance, models are often difficult to interpret and expensive to train, since training requires unrolling sequences of predictions in time.
  As an alternative to end-to-end trained state estimation, we propose a novel particle filtering algorithm in which models are trained from individual state transitions, fully exploiting the Markov property in robotic systems. 
 In this framework, measurement models are learned implicitly by minimizing a denoising score matching objective. 
  At inference, the learned denoiser is used alongside a (learned) dynamics model to approximately solve the Bayesian filtering equation at each time step, effectively guiding predicted states toward the data manifold informed by measurements. 
 We evaluate the proposed method on challenging robotic state estimation tasks in simulation, demonstrating competitive performance compared to tuned end-to-end trained baselines.
Importantly, our method offers the desirable composability of classical filtering algorithms, allowing prior information and external sensor models to be incorporated without retraining.
\end{abstract}

%%%%%%%%%%%%%%%%%%%%%%%%%%%%%%%%%%%%%%%%%%%%%%%%%%%%%%%%%%%%%%%%%%%%%%%%%%%%%%%%
\section{Introduction}

State estimation is a ubiquitous problem in robotics, with applications ranging from robotic in-hand manipulation~\citep{Qi2023-th, Rostel2022-gu} to localization in open environments~\citep{Karkus2018-ur,Koide2024-yo}.
Traditionally, the problem is approached with Bayesian filtering techniques~\citep{thrun2002probabilistic,Doucet2008-eo}, which recursively integrate available measurements $y_t$ and control inputs $u_t$ to compute an estimate of the posterior distribution $p(x_t | y_{1:t}, u_{1:t})$ over (unobserved) states $x_t$ at each timestep $t$.
As dynamics and sensor models are often expensive to compute or even unknown, the problem is increasingly addressed by learning-based methods~\citep{Haarnoja2016-ay,Karkus2018-ur,Jonschkowski2018-go,Rostel2022-gu}.
State estimation is then often treated as a sequence modeling problem, where the goal is to learn a mapping from the history of measurements to the distribution over states.
Accordingly, given a dataset with ground truth states for training, the problem of state estimation can be addressed by supervised methods for sequential data, such as recurrent neural networks (RNNs)~\cite{hochreiter1997long, cho2014learning} or Transformers~\cite{vaswani2017attention, Qi2023-th}.
While such methods are convenient to deploy and often perform well, they lack the modularity and interpretability of Bayesian filters. 
For example, RNNs and Transformers offer no immediate way of initializing to an arbitrary prior $p(x_0)$, as their hidden states are not readily interpretable. 

Differentiable Bayesian Filters~\citep{Haarnoja2016-ay,Karkus2018-ur,Jonschkowski2018-go,kloss2021train} (DFs) aim to reintroduce these properties by combining the algorithmic structure of Bayesian filters with learned models. 
To achieve the desired performance, DFs are trained end-to-end by unrolling the filtering algorithm sequentially and optimizing the model parameters via backpropagation through time (BPTT).
However, this end-to-end training procedure is costly and difficult to scale, especially for Differentiable Particle Filters (DPFs)~\citep{Karkus2018-ur}, which require BPTT for each particle.
Moreover, changing sensor models or including additional information requires retraining.

Another consequence of the end-to-end training paradigm, in DFs, RNNs, and Transformers, is the tendency to overfit to the specific sequential structure of training sequences. 
One remedy would be to exploit the Markov property in dynamical systems by training with single-step objectives on individual transitions rather than full sequences.
However, recursively using models trained with single-step objectives in practice often fails as intermediate predictions eventually leave the data manifold~\citep{Lutter2021-dv} (i.e., the manifold of states seen during training).

\begin{figure}[t]
\centering
\includegraphics[width=\linewidth]{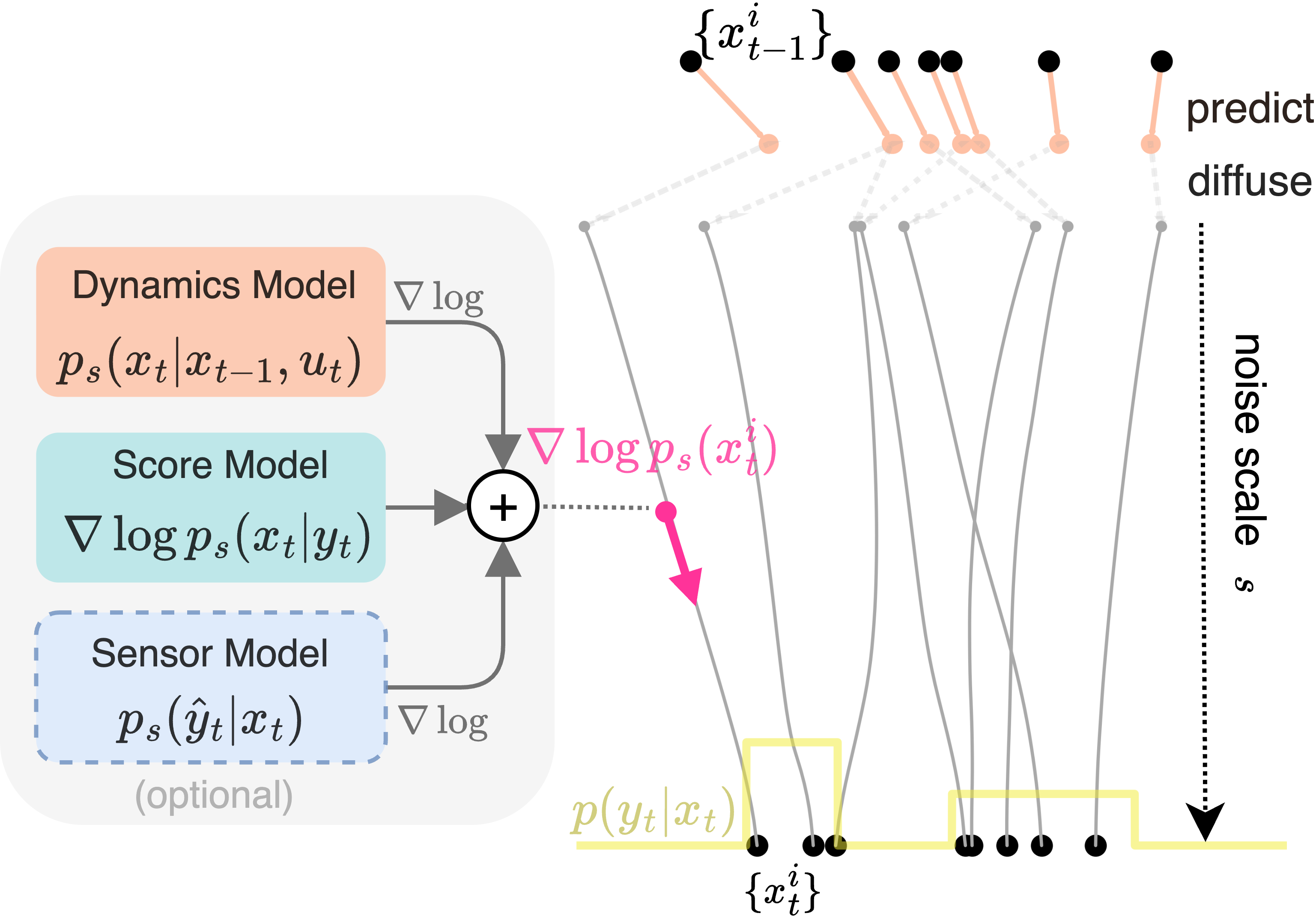}
\caption{One timestep of inference with Denoising Particle Filters (DnPF).
DnPF approximates the posterior $p(x_t|y_{1:t}, u_{1:t})$ as a set of particles $\{x_{t}^{i}\}_{i=1}^N$ and recursively solves the Bayesian filtering equation in score space.
At each timestep $t$, each particle undergoes a series of integration steps $s = 0\rightarrow 1$, moving according to the sum of score terms for dynamics $\nabla\log p_s(x_t|x_{t-1}, u_t)$, data likelihood score $\nabla\log p_s(x_t|y_t)$, and, optionally, a known external sensor model $\nabla\log p_s(\hat{y}_t | x_t)$.
Instead of starting from pure noise at each timestep, particles are \textit{warm-started} with noise-perturbed predictions from the (learned) dynamics model (top).
The data likelihood score $\nabla\log p_s(x_t|y_t)$ is predicted by a \textit{score network} $D(x_t, y_t, s)$, which can be trained efficiently via denoising score matching.
}
\label{fig:intro}
\end{figure}

In this work, we introduce a new learning framework that addresses these challenges. 
Our method trains dynamics and measurement models using single-step objectives and combines them through a diffusion-based particle filtering procedure. 
Posterior sampling is approximated by integrating the ODE underlying diffusion model sampling, where the denoising (measurement) model corrects predictions toward the data manifold at each step. 
Unlike end-to-end training, our approach produces modular models that can flexibly integrate priors or external sensor models without retraining.

\begin{table}[t]
  \centering
  \begin{tabular}{lccccc}
      \toprule
      & Unlimited Context & Training Efficiency & Modularity\\
      \midrule
      RNN & \ding{51} & \ding{55} & \ding{55}  \\
      Transformer & \ding{55} & \ding{51} & \ding{55}  \\
      DPF & \ding{51} & \ding{55} & \ding{51}  \\
            \midrule
      DnPF (ours) & \ding{51} & \ding{51} & \ding{51}\ding{51}\\
      \bottomrule
  \end{tabular}
  %\caption{Comparison of learning-based state estimation methods}
  %\label{tab:comparison}
\end{table}
Our contributions are:
\begin{itemize}
  \item We propose a modular learning framework for state estimation that uses only single-step objectives, avoiding end-to-end training on sequences.
  \item We introduce a novel diffusion-based particle filtering inference scheme for approximating posterior sampling in Bayes filters that mitigates distributional drift. 
  We also propose a likelihood constrained diffusion process to ensure that particles remain close to the data manifold induced by measurements (likelihood manifold).
  \item In state estimation tasks with partial observability, non-linear dynamics, and high-dimensional state spaces, we demonstrate that our approach achieves competitive or superior accuracy to end-to-end methods, while enabling efficient, scalable training.
  \item We show that our models can be flexibly and effectively composed with known sensor models without retraining.
\end{itemize}

\section{Related Work}
\label{sec:related_work}
State estimation is an extensively studied problem in robotics and has been addressed with a variety of methods, both learning-based and non-learning based.
In the regime of non-learning-based approaches, particle filters have been widely used for non-linear Bayesian filtering in robotics~\cite{thrun2002probabilistic}. 
A plethora of PF variants have been proposed to address challenges arising from particle starvation and weight degeneracy~\cite{Doucet2008-eo,van-der-Merwe2000-ot}, particularly in the context of tactile sensing~\cite{koval2017manifold,Wirnshofer2019-en}.
Multiple works have considered particle filtering based on the Stein Score~\citep{pulido2019kernel, Maken2022-qa, Koide2024-yo} via Stein Variational Gradient Descent (SVGD)~\citep{Liu2016-ri}. 
We see the combination of SVGD and learning sensor models via Denoising Score Matching as an interesting direction for future work.

In the regime of learning-based methods, our approach is related to Differentiable Particle Filters (DPF)~\citep{Jonschkowski2018-go,Karkus2018-ur,Rostel2022-gu}, as we combine the particle filtering algorithm with learned models.
However, while DPFs unroll the particle filtering algorithm sequentially and optimize dynamics and measurement models by BPTT, we learn models via single-step objectives and combine them in a denoising scheme at inference.

Diffusion models have been successfully applied to a variety of tasks, including image generation~\citep{ho2020denoising, Song2020-in,Dhariwal2021-rj, ho2022classifier}, imitation learning~\cite{chi2023diffusionpolicy}, and model-based control~\citep{suh2023fighting}. 
\citet{Rozet2023-gu} leverage diffusion models for learning priors over state trajectories and use them for data assimilation in a guided diffusion~\citep{Dhariwal2021-rj, ho2022classifier} process. 
To our knowledge, integrating models learned via score matching into particle filters to approximately solve the Bayesian filtering equation has not been considered before.

\section{Background}

\subsection{Bayesian State Estimation}
\label{sec:bayesian_state_estimation}
Given a dynamical system of which the state at timestep $t$ is fully described by $x_t \in \mathcal{X}$ as well as known control inputs $u_t \in \mathcal{U}$ and measurements $y_t \in \mathcal{Y}$, the goal of filtering is to find the \textit{posterior distribution} $p(x_t | y_{1:t}, u_{1:t})$ over states given measurements and control inputs up to time $t$.
Assuming the system is Markovian with \textit{dynamics model} $p(x_t | x_{t-1}, u_t)$ and \textit{measurement model} $p(y_t | x_t)$, the posterior can be computed recursively as

\begin{align}
  p(x_t | y_{1:t}, u_{1:t}) 
  &= p(y_t | x_t) \int p(x_t | x_{t-1}, u_t) \nonumber \\
  &\quad \cdot p(x_{t-1} | y_{1:t-1}, u_{1:t-1}) \, \mathrm{d} x_{t-1}
  \label{eq:filtering}
\end{align}
where we imply a prior distribution $x_0 \sim p(x_0)$ and a \textit{policy} $u_t \sim p(u_t |y_{1:t})$.
For the case of linear and Gaussian dynamics and measurement models, the posterior can be computed in closed form using the Kalman filter~\cite{kalman1960new}. 
In most other cases, however, \cref{eq:filtering} is intractable, and approximations are required.
In the case of non-linear models and/or highly non-Gaussian posteriors, a common non-parametric approach is the \textit{particle filter} (PF)~\cite{Doucet2008-eo} which represents the posterior by a finite set of weighted samples (particles) with states $x_t^{i}$ and weights $w_t^{i}$ for $i=1,\ldots,N$ and $\sum_{i=1}^N w_t^{i} = 1$.
The posterior is then formally approximated as $ p(x_t | y_{1:t}, u_{1:t}) \approx \sum_{i=1}^N w_t^{i} \delta(x_t - x_t^{i})$ with the Dirac delta function $\delta$.

\subsection{Diffusion Models}
\textit{Diffusion}~\citep{sohl2015deep, ho2020denoising, Song2019-jk, Song2020-in} or \textit{flow} models~\cite{Lipman2022-bq} are a class of generative models that sample from a target distribution $p(x)$ by iteratively denoising samples from a known initial distribution $p_{\text{init}}(x)$.
For this, an interpolating sequence of distributions $p_s(x)$ for $0 \leq s\leq 1$ is defined such that $p_0(x) = p_{\text{init}}(x)$ and $p_1(x)=p(x)$.
Sampling from the target distribution is then performed by first sampling\footnote{In this section only, perturbed variables for noise scale $s$ are denoted as $x_s$, which is distinct from indexing physical time $t$ as in $x_t$ as in the rest of this paper.} $x_0\sim p_0$ and numerically integrating the ordinary differential equation (ODE)
\begin{align}
  \label{eq:ode}
  \frac{\mathrm{d}}{\mathrm{d}s}x_s = v(x_s, s)
\end{align}
 where the vector field $v:\mathbb{R}^n\times [0, 1] \rightarrow\mathbb{R}^n$ is constructed such that $x_s \sim p_s$ for all $s$.
Choosing Gaussian perturbation kernels $p_s(x_s|x_1) = \mathcal{N}(x_s; \alpha_s x_1, \beta_s^2 \mathrm{I})$ with monotonic noise scale sequences $\alpha_s, \beta_s \in [0,1]$ such that $\alpha_0 = \beta_1 =0$ and $\alpha_1 = \beta_0 = 1$ allows for a closed-form conversion between $v$ and the score $\nabla_{x_s} \log p_s(x_s)$ of the (marginal) perturbed distribution as \cite{Holderrieth2025-oy}
\begin{align}
v(x_s, s) = \left(\beta_s^2\frac{\dot{\alpha}_s}{\alpha_s} -\dot{\beta}_s\beta_s\right) \nabla_{x_s} \log p_s(x_s) + \frac{\dot{\alpha}_s}{\alpha_s} x_s.
\end{align}
A key finding in the diffusion literature is that tractable and efficient objectives for learning approximations to the vector field $v(x_s, s)$ or score $\nabla_{x_s} \log p_s(x_s)$ are available. 

In particular, \citet{ho2020denoising} propose to learn a \textit{noise model} $D(x_s, s)$ that predicts the noise $\epsilon \sim \mathcal{N}(0, \mathrm{I})$ in the perturbed sample $x_s = \alpha_s x + \beta_s \epsilon$ by minimizing the \textit{Denoising Score Matching} objective
\begin{align}
  \label{eq:ho_simple}
 \mathbb{E}_{x, s\sim[0,1], \epsilon \sim \mathcal{N}(0, \mathrm{I})} \left[ \left\| \epsilon - D\left(x_s, s\right) \right\|^2 \right].
\end{align}
where the expectation is taken over samples $x\sim p(x)$ from the dataset.
As shown by \citet{Song2020-in}, the output of the optimal noise model $D^{*}$ trained with objective \cref{eq:ho_simple} is related to the score as
\begin{align}
  \label{eq:noise_term}
   \nabla_{x_s} \log p_s(x_s) = -\frac{ D^{*}(x_s, s)}{\beta_s}.
\end{align}
In the rest of this paper, to avoid cluttered notation when indexing both physical time $t$ and noise scale $s$ in states $x_{t,s}$, whenever unambiguous we drop the noise scale index on variables and denote distributions over perturbed data as $p_s(x_t)$.
\section{State Estimation with Denoising Particle Filters}
\label{sec:dnf}
At its core, our approach is based on sampling from the posterior $p(x_t | y_{1:t}, u_{1:t})$ by solving the ODE \cref{eq:ode}, which requires computing its score $\nabla_{x_t} \log p_s(x_t | y_{1:t}, u_{1:t})$ for multiple noise scales $s$.
Taking the gradient w.r.t. $x_t$ of the logarithm of the recursive filtering equation \cref{eq:filtering} and approximating the integral with a set of particles $\{x_{t-1}^{i}\}$ representing the posterior at the previous timestep, we obtain

\begin{align}
  \label{eq:filter_score}
  \nabla_{x_t} \log p_s(x_t | y_{1:t}, u_{1:t}) 
  &= \nabla_{x_t}\log p_s(y_t | x_{t}) \nonumber \\
  & + \nabla_{x_t} \log \sum_i  p_s(x_t | x_{t-1}^{i}, u_t).
\end{align}

Eq.~\cref{eq:filter_score} makes sampling from the filtering equation tractable for arbitrary target distributions, assuming that the score terms $\nabla_{x_t}\log p_s(y_t | x_{t})$ and $\nabla_{x_t} \log \sum_i  p_s(x_t | x_{t-1}^{i}, u_t)$ can be evaluated.
As samples are drawn directly from the posterior, the resulting particle filter does not require importance sampling.
From the perspective of diffusion models, we derive a guided diffusion process such that the target distribution approximates $p(x_t | y_{1:t}, u_{1:t})$.
In the remainder of this section, we first discuss learning objectives for the score terms.
Then, we describe the combination of the (learned) models at inference in a particle filter.

\subsection{Training Objectives}
For the measurement likelihood score, Bayes' rule yields $\nabla_{x_t} \log p(y_t | x_t) = \nabla_{x_t} \log p(x_t | y_t) - \nabla_{x_t} \log p(x_t)$.
Since it is often easier to learn a generative model for $p(x_t | y_t)$ than for $p(y_t | x_{t})$ (e.g., consider the case in which $\mathcal{Y}$ is the space of images), we propose to learn a model approximating $\nabla \log p_s(x_t | y_t)$ and $\nabla \log p_s(x_t)$ by denoising score matching.
Specifically, we learn an observation-conditioned denoising model $D(x_t, y_t, s)$ by minimizing the objective

\begin{align}
  \mathcal{L}_{\text{lh}} = \mathbb{E}_{(x_t, y_t), s, \epsilon\sim\mathcal{N}(0, \mathrm{I})} \left[ \left\| \epsilon - D(x_{t,s}, y_t, s) \right\|^2 \right]
\end{align}

and recover the score as in \cref{eq:noise_term}.
In practice, the same model can be used to predict the score of the prior $\nabla \log p_s(x_t)$  by sporadically dropping the conditioning during training~\citep{ho2022classifier}.
Analogous to classifier-free guidance~\citep{ho2022classifier}, we then perform denoising steps with respect to 

\begin{align}
  \label{eq:llh_score}
  \epsilon_{\text{lh}} = -\beta_s\left[(1 + \eta) \nabla_{x} \log p_s(x_t | y_t) - \eta\nabla \log p_s(x_t)\right]
\end{align}
with \textit{guidance strength} $\eta \geq 0$.
Substituting this into the original filtering equation reveals that this approximation introduces a bias of $p(y_t | x_t)^{\eta}p(x_t)$ at each timestep.
Although this bias could be corrected via importance sampling, we do not correct for it in practice. 
In the context of recursive prediction with learned models, we find that the bias on $p(x_t)$ is actually beneficial, as it mitigates distributional drift during inference (see \cref{sec:ablation}).

The dynamics score term is a mixture of the scores of the (perturbed) dynamics model $\nabla_{x_t} \log p_s(x_t | x_{t-1}^{i}, u_t)$, weighted by the relative prior of each ancestor. 
In our implementation we let each particle evolve from a single ancestor $x_{t-1}^{i}$, which is a common approximation made in particle filtering \citep{Doucet2008-eo}.
An alternative would be evaluating the full mixture prior~\cite{Klaas2012-nk}, which we leave for future work.

There are multiple options for obtaining the score of the dynamics model $\nabla_{x_t} \log p_s(x_t | x_{t-1}^{i}, u_t)$. 
One option is to approximate the score by learning a denoising model conditioned on $x_{t-1}$ and $u_t$ with objective \cref{eq:ho_simple}. 
Alternatively, if $p_s(x_t | x_{t-1}, u_t)$ is known, its score could be evaluated analytically (or numerically).
In this work, we choose a third variant and learn a parametric dynamics model $f$ by maximizing the log-likelihood of single-step predictions

\begin{align}
  \mathcal{L}_{\text{dy}} = \mathbb{E}_{(x_t, x_{t-1}, u_t)} \left[ \log \mathcal{N}\left(x_t; \mu_f, \Sigma_f \right) \right]
\end{align}
where the mean $\mu_f$ and covariance $\Sigma_f$ of the gaussian are predicted by two heads $f_{\mu}$, $f_{\sigma}$ of a feedforward network:

\begin{align}
\label{eq:dyn_mean}
\mu_f = x_{t-1} + f_{\mu}(x_{t-1}, u_t)\\
\label{eq:dyn_sigma}
\Sigma_f = \exp(f_{\sigma}(x_{t-1}, u_t)) \mathrm{I}
\end{align}

The benefit of parametrizing $p(x_t | x_{t-1}, u_t)$ as Gaussian in the context of DnPF is that the score of the perturbed transition can be evaluated in closed form in each denoising step as 
\begin{align}
  \label{eq:dyn_noise}
\epsilon_{\text{dy}}= - \beta_s \nabla_{x_t} \log p_s(x_t | x_{t-1}, u_t) = \beta_s  \Sigma_s^{-1}(\mu_s - x_t)
\end{align}
with $\mu_s = \alpha_s\mu_f$ and $\Sigma_s = \alpha_s^2 \Sigma_f + \beta_s^2 \mathrm{I}$.
This way, the learned model $f$ has to be evaluated only once per particle and timestep, as opposed to each denoising step.

\subsection{Inference Procedure}
Given a set of particles $\{x_{t-1}^{i}| i=1,..., N\}$ representing the posterior at the previous timestep $p(x_{t-1} | y_{1:t-1}, u_{1:t-1})$, \cref{eq:filter_score} gives an opportunity to sample from the posterior at the next timestep:  
Starting from pure noise samples $x_{t,0}^{i} \sim \mathcal{N}(0, \mathrm{I})$, each particle evolves independently through a sequence of denoising steps $s=0, \ldots, 1$ according to the predicted noise 

\begin{align}
  \label{eq:dnf_noise}
  \epsilon^{i} =  \epsilon^{i}_{\text{lh}} + \epsilon^{i}_{\text{dy}},
\end{align}
where $\epsilon^{i}_{\text{lh}}$ (see \cref{eq:llh_score}) accounts for the learned measurement likelihood and $\epsilon^{i}_{\text{dy}}$ (see \cref{eq:dyn_noise}) accounts for the dynamics prior.

As sampling with diffusion models for many denoising steps $S= 1/\Delta s$ can be expensive, we warm-start the denoising process as follows:
For each previous particle $x_{t-1}^{i}$, we first predict the next state $\hat{x}_{t}^{i} = x_{t-1}^{i} + f_{\mu}(x_{t-1}^{i}, u_t)$ using the (learned) dynamics model. 
Then, for a chosen \textit{warm start noise scale} $s_{\mathrm{w}}\in [0, 1)$ we sample a perturbed initial state $x_{t,s_{\mathrm{w}}}^{i} \sim \mathcal{N}(\alpha_{s_{\mathrm{w}}} \hat{x}_{t}^{i}, \beta_{s_{\mathrm{w}}}^2 \mathrm{I})$, i.e., we interpolate between pure noise and the dynamics model prediction.
For each denoising step $s=s_\mathrm{w}, \ldots, 1$, we then compute the predicted noise using \cref{eq:dnf_noise}.
In practice, we find that values of $s_{\mathrm{w}}\in [0.5, 0.9]$ with a total of 5 to 25 denoising steps work well for the studied tasks.

We summarize the inference procedure in \cref{alg:dnf}.

\subsection{Likelihood-Constrained Diffusion}
\label{sec:constraint}
A consequence of approximating the posterior with a finite set of samples is that the supports of the prior (dynamics) distribution and the likelihood function may be disjoint.
This is particularly common when the dynamics model is nearly deterministic, in which case even optimally solving the recursive Bayes filter equation \cref{eq:filtering} may fail to yield a well-approximated posterior. 
To address this, we devise a constrained optimization procedure ensuring that particles $x_t^{i}$ remain close to the likelihood manifold $p(x_t | y_t)$, even if unsupported by the prior $p(x_t | x_{t-1}^{i}, u_t)$.

We use the magnitude of the noise model output $D(x_t^{i}, y_t, s)$ at each denoising step as a proxy for the distance to the data manifold induced by $p_s(x_t | y_t)$. This can be justified by interpreting denoising score matching as maximizing the Evidence Lower Bound (ELBO)~\citep{ho2020denoising}.  
This motivates us to define a cost function
\begin{align}
  \label{eq:constraint_cost}
  c(\epsilon) = \left|\epsilon\right| - \theta
\end{align}
with \textit{magnitude threshold} $\theta > 0$. To enforce the inequality constraint $c(\epsilon) \leq 0$ during denoising, we employ an augmented Lagrangian scheme, replacing \cref{eq:dnf_noise} with
\begin{align}
  \label{eq:constraint}
  \epsilon &= \epsilon_{\text{lh}} + \frac{1}{1 + \lambda + \rho c(\epsilon_{\text{lh}})_{+}} \epsilon_{\text{dy}} \\
  \lambda & \leftarrow \left(\lambda + \rho c(\epsilon_{\text{lh}})\right)_{+}
\end{align}
where $\lambda\geq 0$ is a Lagrange multiplier, $\rho \geq 0$ is a penalty parameter and $c_{+} = \max(0, c)$.
Equation \cref{eq:constraint} effectively scales down the dynamics term $\epsilon_{\text{dy}}$ if the distance to the data manifold is larger than $\theta$.
In practice, we find it beneficial to keep per-dimension costs and Lagrange multipliers, independently scaling the noise terms for each dimension of the state space with a global scalar $\theta$.
Example rollouts comparing DnPF with and without the likelihood constraint are shown in the experiments section (\cref{fig:spin_plot}).
\begin{algorithm}[t]
  \caption{Denoising Particle Filter (DnPF)}
  \begin{algorithmic}[0]
  
  \If{prior $p(x_0)$ available} 
  \State initialize $x_0^{i} \sim p(x_0)$
  \Else
  \State initialize $x_0^{i} \sim p(\cdot|y_0)$ using $D(\cdot | y_0)$
  \EndIf
  \For{each timestep $t = 1, \ldots $}
  \For{each particle $i = 1, \ldots, N $}
      \State Predict $\hat{x}_t^{i}, \Sigma_t^{i}$ using \cref{eq:dyn_mean}, \cref{eq:dyn_sigma}
      \State $x_{t}^{i} = \alpha_{s_{\mathrm{w}}} \hat{x}_t^{i} + \beta_{s_{\mathrm{w}}}\epsilon$, \quad $\epsilon\sim\mathcal{N}(0, \mathrm{I})$
      \For{each denoising step $s = s_{\mathrm{w}}\ldots 1$ }
          \State $\epsilon_{\text{lh}}^{i} = (1 + \eta) D(x_{t,s}^{i}, y_t, s) - \eta D(x_{t,s}^{i}, s)$ 
          \State $\epsilon_{\text{dy}}^{i} = -\beta_s \nabla \log p_s(x_{t}^{i}| \hat{x}_t^{i} ,\Sigma_t^{i})$ 
          \State $\epsilon^{i} = \epsilon_{\text{lh}}^{i} + \epsilon_{\text{dy}}^{i}$
          \State $x_{t}^{i} \leftarrow \textsc{Denoise}(x_{t}^{i}, \epsilon^{i}, s)$
        \EndFor
      \EndFor
  \EndFor
  \end{algorithmic}
  \label{alg:dnf}
\end{algorithm}

\subsection{Implementation}
\begin{figure}[t]
\includegraphics[width=\linewidth]{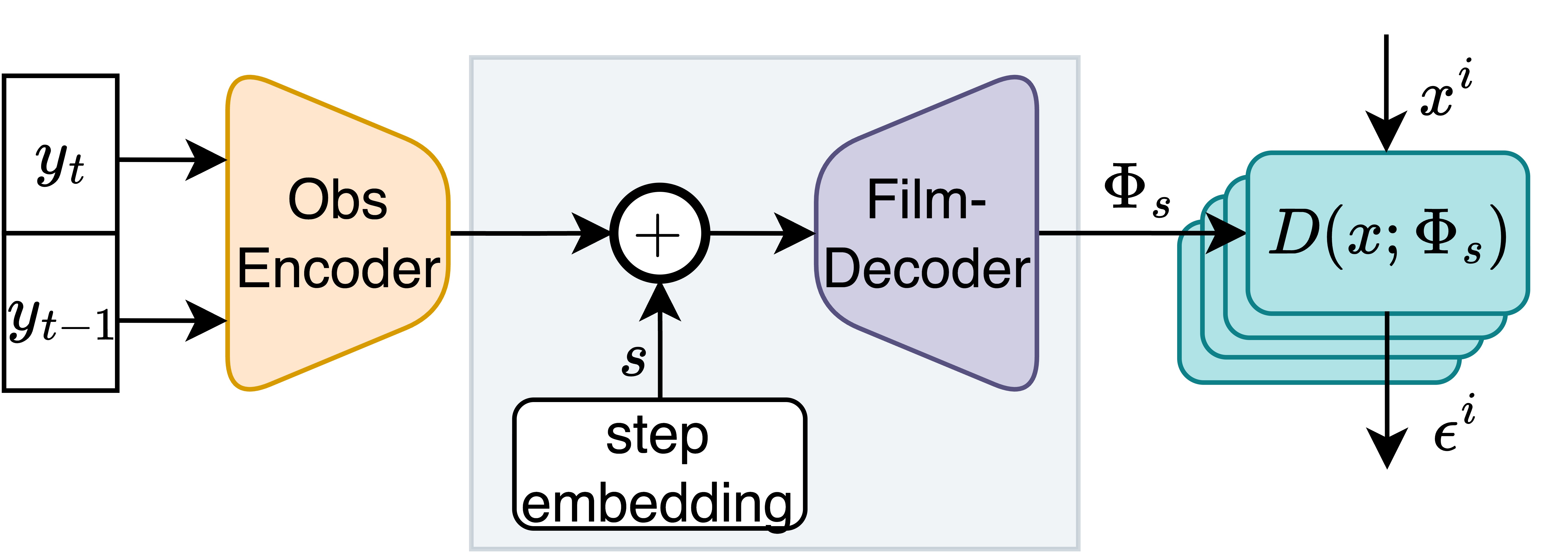}
\caption{DnPF network structure for efficient denoising inference.}
\label{fig:network_structure}
\end{figure}

In our implementation of DnPF, we make several design decisions to enable fast inference.
First, since each particle $x_t^{i}$ is independent at each denoising step, we can parallelize over particles on the GPU.
Second, we compute a shared observation encoding $y_{\text{enc}} = E(y_t, y_{t-1})$ once per timestep using a neural network $E$.
The observation encoder $E$ is parametrized as a feedforward network for low-dimensional observations and as a convolutional network for image observations.
We additionally condition on the previous observation $y_{t-1}$ to ease the prediction of velocity components in the state space.
For a fixed number of denoising steps $s=s_\mathrm{w},\ldots,1$, we precompute FiLM conditioning~\cite{perez2018film} vectors $\Phi_{s_\mathrm{w}} \dots \Phi_1$ as $\Phi_s = F(y_{\text{enc}}, s)$ in parallel, where $F$ is a feedforward neural network.
The number of conditioning vectors is independent of the number of particles $N$ and can be reused across all particles in a denoising step.
The FiLM vectors $\Phi_s$ condition the denoising model $D(x_t^{i}; \Phi_s)$, which can be kept small to accelerate sequential diffusion inference.
We give an overview of the denoising architecture in \cref{fig:network_structure}.

For all experiments, we use 4-layer feedforward networks with layer normalization~\citep{ba2016layer}, skip connections and inverted bottlenecks~\citep{vaswani2017attention}, with 256 hidden units per layer for $F$, and the dynamics model $f$, and 128 hidden units for the denoising model $D$.
For diffusion inference, we use the DDIM schedule~\cite{Song2020-er}, parametrizing $\alpha_s = \sqrt{\bar{\alpha}_s}$ and $\beta_s = \sqrt{1 - \bar{\alpha}_s}$ with $\bar{\alpha}_s \in [0, 1]$, and solve \cref{eq:ode} via Euler integration.
Note that our algorithm is not restricted to DDIM and is compatible with any diffusion or flow model sampling scheme, provided score terms can be computed.
DnPF introduces several inference hyperparameters: the number of denoising steps $S$, warm start noise scale $s_{\mathrm{w}}$, guidance strength $\eta$, and constraint threshold $\theta$. 
Note that the DnPF training is agnostic to these inference parameters.
While DnPF behaves well for reasonable ranges of these parameters, we find it beneficial to sweep over inference parameters for maximizing performance with respect to the chosen evaluation metric after training.
See \cref{sec:runtime} for an analysis of runtime-performance tradeoffs.

\section{Experiments}
\label{sec:experiments}
In this section, we evaluate the Denoising Particle Filter (DnPF) proposed in \cref{sec:dnf} on a set of challenging state estimation tasks in simulation.
We first describe the experimental setup, including the tasks (\cref{sec:tasks}), evaluation metrics (\cref{sec:metric}) and considered baselines (\cref{sec:baselines}).
Then, we analyze the performance of DnPF on the tasks (\cref{sec:results}). 
Finally, we show the modularity of DnPF by integrating external sensor models post-training (\cref{sec:sensor_fusion}).

\subsection{Task Setup}
\label{sec:tasks}
\begin{figure}[t]
\centering
\includegraphics[width=.9\linewidth]{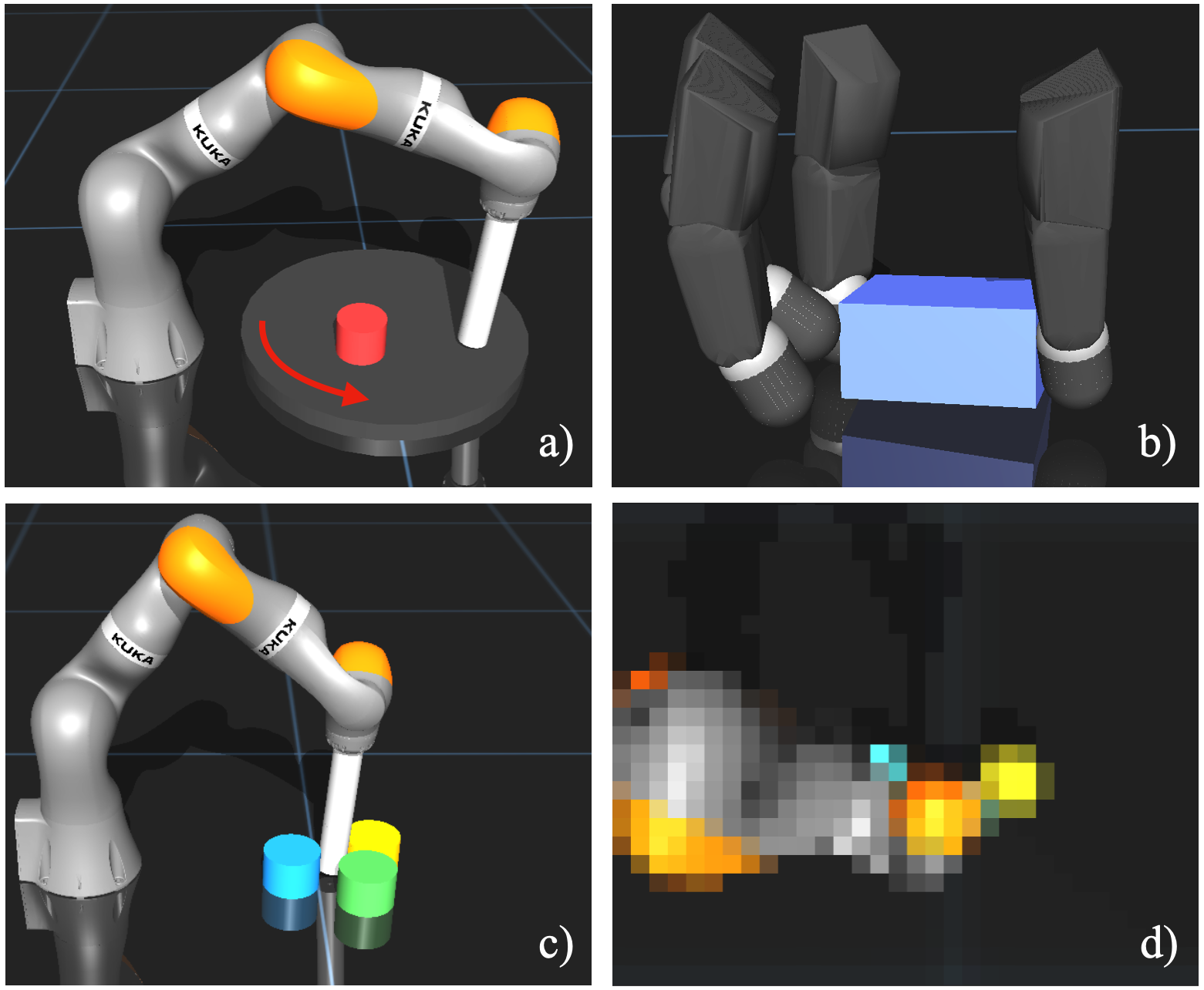}
\caption{Simulated state estimation tasks used in the experiments. \textbf{a)} In the \textit{Manipulator Spin} task a 7-dof manipulator interacts with an object on a spinning table. \textbf{b) }\textit{Multi-fingered Manipulation}: a 12-dof robotic hand manipulates a rigid object.
\textbf{c)} \textit{Cluttered Push} a manipulator pushes around 3 cylindrical objects on a plane. \textbf{d)} Image observation showing a top-down view with occlusions of the \textit{Cluttered Push} task.}
\label{fig:tasks_overview}
\end{figure}

We evaluate our approach on three simulated state estimation tasks shown in \cref{fig:tasks_overview}. 
In the \textbf{Manipulator Spin} task, a 7-DoF compliantly controlled manipulator interacts with an object that follows a circular trajectory on a spinning table. 
Observations include only the manipulator joint angles and the difference from the applied control inputs $y \in \mathbb{R}^{14}$ (allowing inference of applied torques). 
We evaluate prediction quality against the ground truth position of the spinning object, which is not directly observable and must be inferred through contact.
Second, in the \textbf{Cluttered Push} task, the same manipulator pushes three cylindrical objects randomly placed on a table using randomly sampled control inputs. 
For this task, observations additionally include a top-down image of the scene, where objects are frequently occluded ($y \in \mathbb{R}^{14} \times [0,1]^{32\times32\times3}$). 
We also black out the image observation with 80\% probability to further increase difficulty.
Lastly, we consider a \textbf{Multi-fingered Manipulation} task where a 12-DoF robotic hand manipulates a rigid object on a flat surface.
Randomly sampled control inputs applied to the finger joints induce object movement via contact.
The only available measurements are the finger joint angles and the difference from the applied control inputs ($y \in \mathbb{R}^{24}$); the goal is to estimate the object's pose in $\mathrm{SE}(3)$.
To predict rotation components, we use a continuous 6D representation~\cite{zhou2019continuity}, mapping values back to $\mathrm{SO}(3)$ in each step.

We implement all tasks in the Mujoco physics engine~\citep{todorov2012mujoco}.
To reflect real-world model uncertainty, we simulate noise in joint measurements and friction. 
For the hand task, we additionally randomize contact surface friction and control gains, and include systematic measurement biases in joint angles across sequences.

While we focus on simulation in this paper to ease quantitative comparisons, training in such randomized simulations would also allow transfer to real-world settings, as shown in prior work on learned state estimation~\cite{Rostel2022-gu}.

\subsection{Evaluation Metric}
\label{sec:metric}
To account for the non-Gaussian nature of the target distributions, we evaluate the fit of the predicted particles using the negative log-likelihood of the ground truth $x$ under a corresponding Gaussian mixture 
\begin{align}
  \label{eq:metric}
  M = -\frac{1}{T}\sum_{t=1}^T \log \sum_{i=1}^N w_t^{i} \mathcal{N}\left(x_t; x_t^{i}, \Sigma\right),
\end{align}  
where the component means are the particle states $x_t^{i}$. 
\cref{eq:metric} approximates the target distribution log-likelihood in the limit $N\to\infty$ and $\Sigma\to 0$.
In practice, we normalize all state dimensions independently and set $\Sigma = \exp(-3)\mathrm{I}$. 
For each task, we collect 10k rollouts of 50 timesteps for training, and 500 rollouts of 100 timesteps for evaluation and testing, respectively.
To robustly compare performance, we report the interquartile mean $M_{\text{IQM}}$ over test sequences.
We report all results in normalized units and divide \cref{eq:metric} by the state space dimensionality $|\mathcal{S}|$ to better facilitate comparison across tasks.
We use a fixed number of 100 particles for all methods and experiments. 
Unless noted otherwise, we provide no prior state information at the first timestep (which would be possible only with DPF and DnPF), resulting in challenging state estimation tasks with high initial uncertainty.

\subsection{Baselines}
\label{sec:baselines}
We consider several competitive baselines for state estimation in dynamical systems, all trained end-to-end to maximize \cref{eq:metric}.
The Differentiable Particle Filter (\textbf{DPF})~\citep{Jonschkowski2018-go,Karkus2018-ur} unrolls the particle filtering algorithm sequentially and optimizes transition and measurement models via BPTT.
\textbf{D2P2F}~\citep{Rostel2022-gu} is an extension of this paradigm that includes observations in the particle proposal.
We also consider two baselines based on GRU~\citep{cho2014learning} and Transformer~\citep{vaswani2017attention} backbones, respectively, for encoding the history of measurements and control inputs.
To predict a set of particles, we learn a particle decoder (PD) that samples particle states $x^{i}$ from the latent embedding end-to-end.
This enables all baselines to represent multi-modal posterior distributions.
We refer to these baselines as \textbf{RNN-PD} and \textbf{Transformer-PD}, respectively. 
For all baselines and DnPF, we train until convergence with early stopping based on validation performance. 
For inference, we use an exponential moving average of training model parameters, which we found to improve performance for all methods. 

\subsection{Results}
\label{sec:results}
\begin{table}
\centering
  \begin{tabular}{@{}l|cccc@{}}
  %\toprule
   & \makecell{\textbf{Manipulator} \\ \textbf{Spin}} &  \makecell{\textbf{Multi-fingered} \\ \textbf{Manipulation}} &  \multicolumn{2}{c}{\makecell{\textbf{Cluttered Push} \\ $\text{ID} \quad \quad \text{OOD}$}} \\
  \midrule
  Transformer &  -0.3  & 4.3 & 0.8 & 9.0\\
  RNN         & 0.1 & \textbf{3.1}   & \textbf{0.3} & 4.4\\
  D2P2F       & -0.8  & 6.2   & 0.7 & 4.7\\
  DPF         & 0.2  & 17.7  & 13.7 & 13.2\\
  \hline 
  DnPF (ours)     & \textbf{-1.1} & \textbf{3.1}   & \textbf{0.3} & \textbf{2.8}\\
  %\bottomrule
  \end{tabular}
  \caption{Evaluation of $M_{\text{IQM}}$ for different models on holdout test sequences (lower is better).}
  \label{tab:results}
  \vspace{-6pt}
\end{table}

\begin{figure}[t]
\centering
\includegraphics[width=\linewidth]{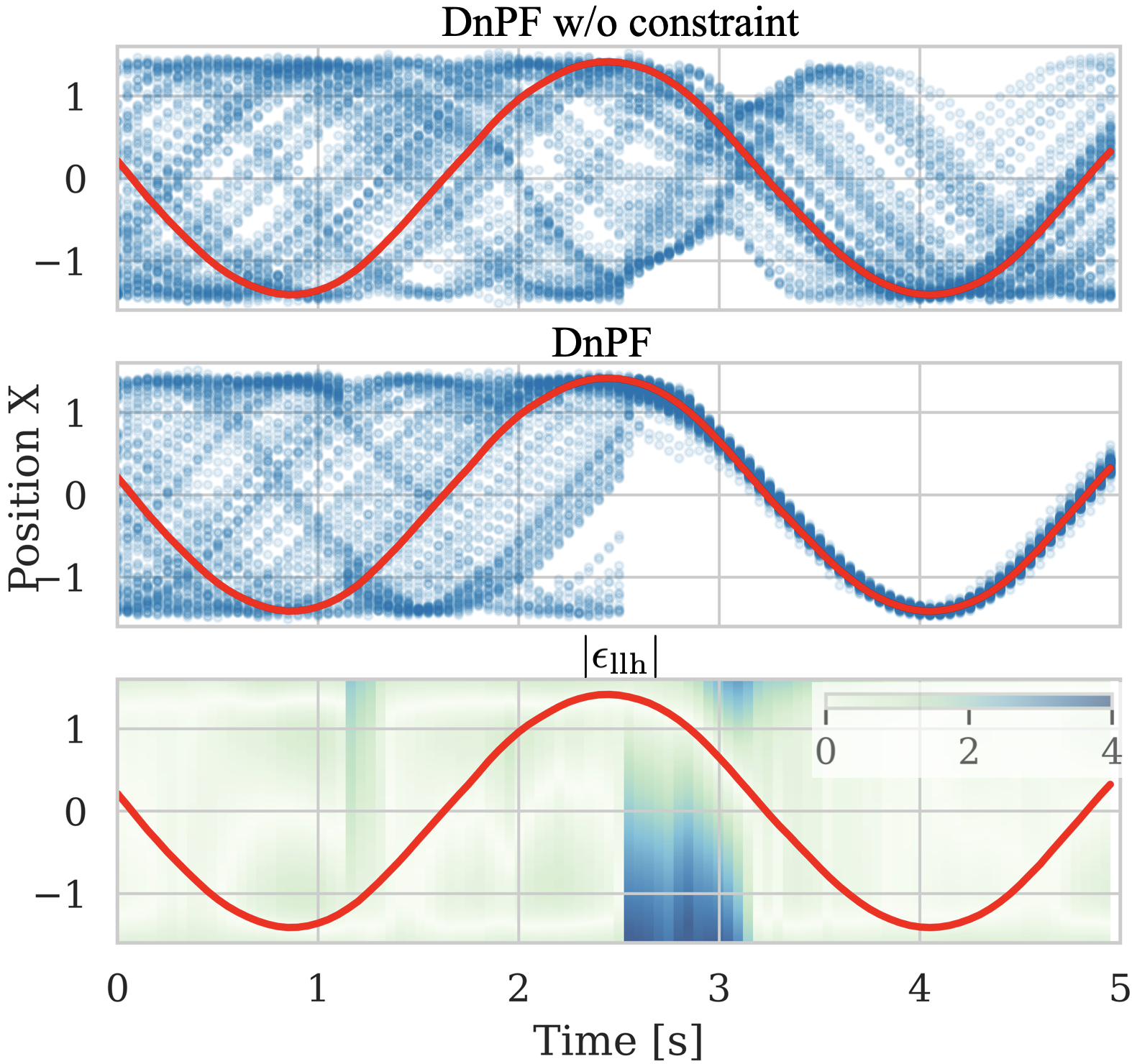}
\caption{DnPF predictions for the \textit{Manipulator Spin} task. Shown are estimates for the normalized X-component of the object position, ground truth shown in red.
At $\sim 2.5\text{s}$, the manipulator end-effector contacts the object, greatly reducing the space of possible object configurations. 
After being pulled toward the induced measurement likelihood, DnPF particles follow the (nearly deterministic) dynamics model again, tracking the ground truth closely.
\textbf{Top}: Particle rollout \textit{without} likelihood constraint, \textbf{Middle:} rollout \textit{with} likelihood constraint ($\theta = 2.0$). \textbf{Bottom:} Magnitude of predicted measurement score $|\epsilon_{\text{lh}}|$.
}
\label{fig:spin_plot}
\end{figure}

\begin{figure*}[h]
  \includegraphics[width=\textwidth]{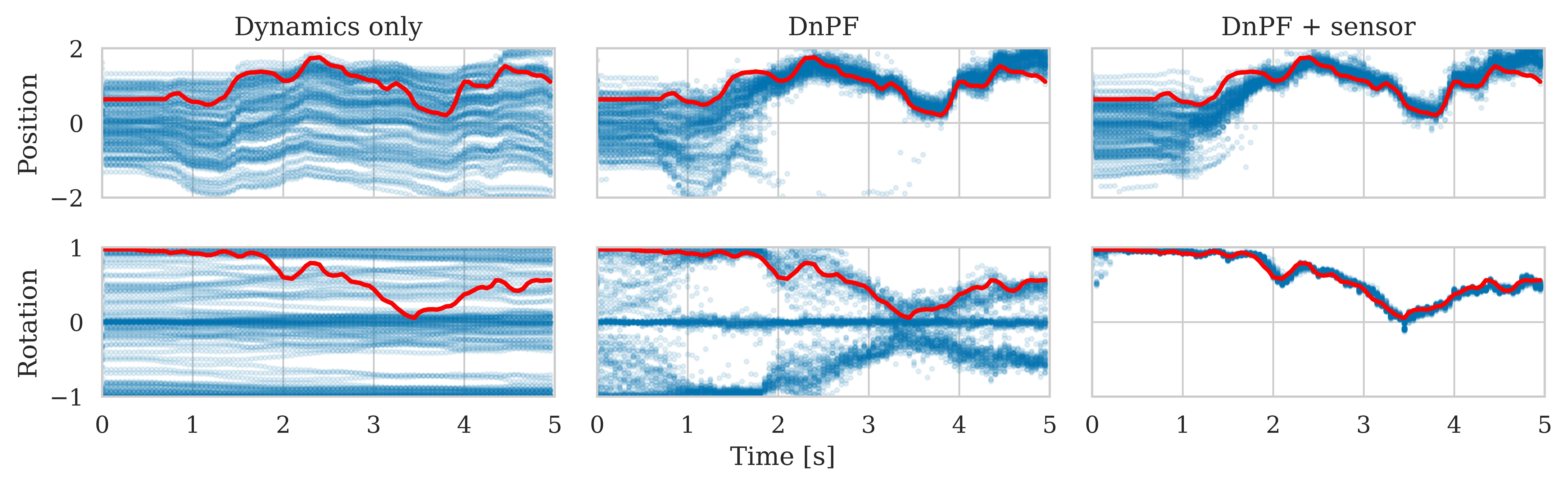}
  \caption{Particle predictions in the \textit{Multi-fingered Manipulation} task, for one normalized position component (top row) and orientation component (bottom row) respectively. 
  Ground truth is shown in red.
  The first column shows particles of an open-loop rollout using only the learned dynamics model $f$.
  The second column shows a particle rollout of DnPF using only proprioceptive measurements. The particle distribution for the rotation component is multi-modal due to the rotational symmetry of the object. 
  In the last column, a simulated external sensor model is additionally included in the DnPF updates, providing (noisy) measurement of rotation $\hat{y}_{\text{rot}} = X_{\text{rot}} + \epsilon$ with $\epsilon \sim \mathcal{N}(0, 0.1^2 \mathrm{I})$.
  }
  \label{fig:prediction_plot}
\end{figure*}

In \cref{tab:results}, we report the performance of DnPF and baselines on the three state estimation tasks described in \cref{sec:tasks}.
On all tasks, DnPF achieves performance on par with or superior to the best end-to-end trained baselines.
We find this encouraging, as the baselines are trained directly to minimize the reported metric \cref{eq:metric}, whereas DnPF is not.
For the \textit{Cluttered Push} task, we evaluate two scenarios: one where image observations are available in 20\% of timesteps, matching the training set (ID), and one where images are available only 2\% of the time (OOD). 
In the OOD case, DnPF substantially outperforms all baselines, indicating that DnPF relies less on the specific training distribution of sequences. 
Qualitatively, we observe that DnPF is able to capture highly multi-modal distributions, as shown in the particle rollouts for the \textit{Multi-fingered Manipulation} task in \cref{fig:prediction_plot}.

\subsection{Score Space Sensor Fusion}
\label{sec:sensor_fusion}

\begin{figure}[t]
\centering
\includegraphics[width=\linewidth]{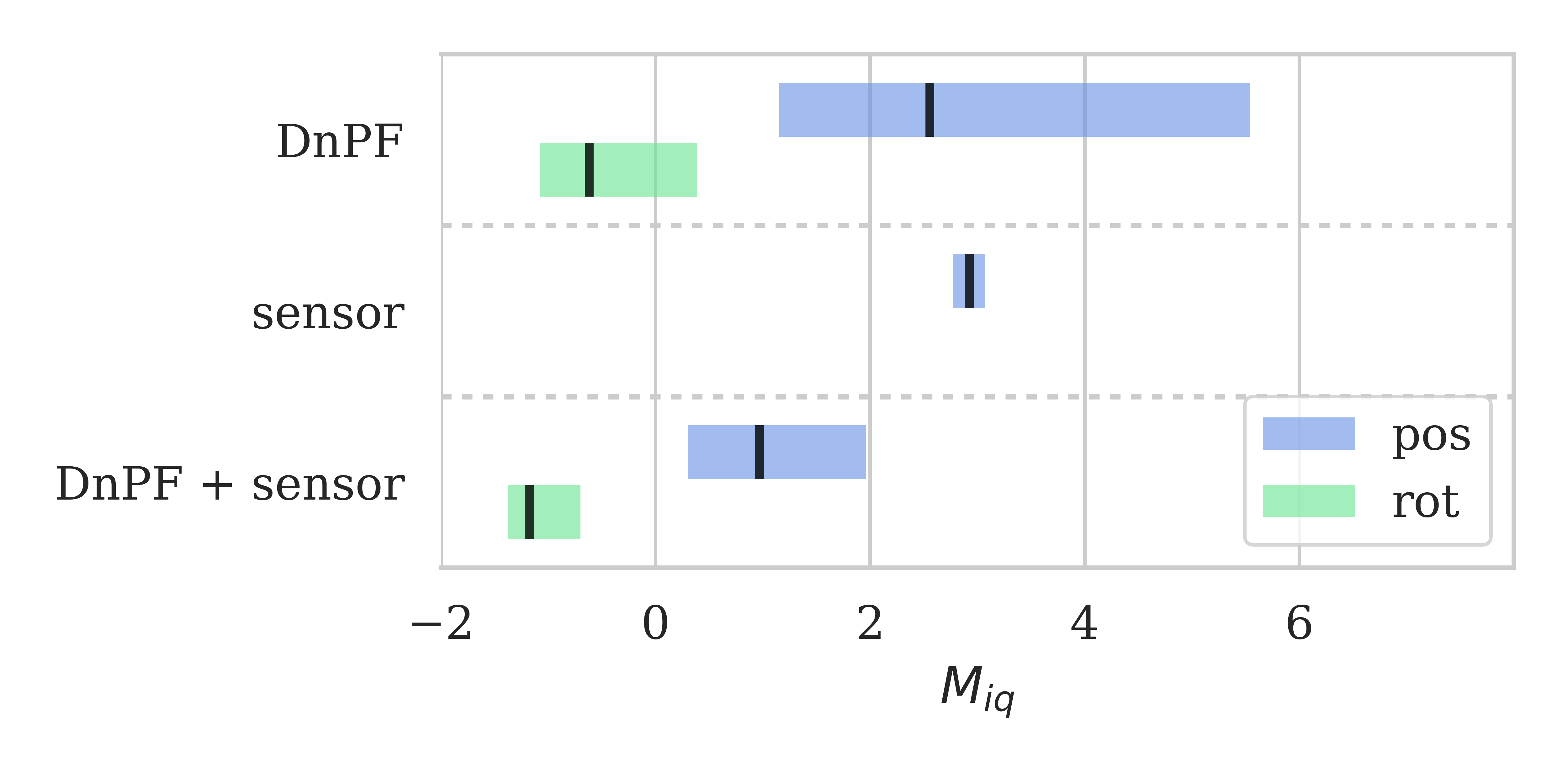}
\caption{Analysis of sensor fusion in DnPF on the Multi-fingered Manipulation task, showing interquartile ranges $M_{\text{IQ}}$ for object position and rotation components, respectively. 
\textbf{DnPF:} with proprioceptive measurements only.
\textbf{sensor:} NLL under the sensor model $p(x|\hat{y}_{\text{pos}})$ measuring only the object position $\hat{y}_{\text{pos}} = X_{\text{pos}} + \epsilon$ with $\epsilon \sim \mathcal{N}(0, 0.5^2 \mathrm{I})$.
\textbf{DnPF + sensor:} DnPF fusing proprioceptive measurements with the additional sensor model $p(\hat{y}_{\text{pos}}|x)$.
}
\label{fig:sensor_plot}
\end{figure}

A major advantage of the score-based DnPF formulation is that it allows for the integration of existing sensor models $p(\hat{y}_t|x_t)$ without retraining. 
To see this, consider the joint measurement likelihood $p(y_{t}, \hat{y}_t | x_t)$ where $\hat{y}_t$ denotes the additional external sensor measurements. 
Assuming conditional independence of $y_t$ and $\hat{y}_t$ given $x_t$, the score of the joint measurement likelihood decomposes as 
\begin{align}
  \nabla_{x_t} \log p(y_{t}, \hat{y}_t | x_t) &= \nabla_{x_t} \log p(y_{t} | x_t) \nonumber \\ &\quad + \nabla_{x_t} \log p(\hat{y}_t | x_t).
\end{align}
Thus, assuming the noise-convolved score $\nabla_{x_t} \log p_s(\hat{y}_t | x_t)$ can be evaluated (straightforward for Gaussian sensor models), we can fuse the additional information $\hat{y}_t$ by adding the score to the denoising step $\epsilon^{i}_{\text{lh}}$ in \cref{eq:dnf_noise}.
We demonstrate this in \cref{fig:sensor_plot} by simulating a noisy external sensor directly measuring object position as $\hat{y}_{\text{pos}} = X_{\text{pos}} + \epsilon$, with $\epsilon \sim \mathcal{N}(0, 0.5^2 \mathrm{I})$. 
As expected, when considering the prediction with respect to the position component, fusing the external sensor model in the DnPF leads to a significant improvement over the cases where either only proprioceptive measurements are available (i.e., the original setting), or only the sensor model is used for prediction.
Interestingly, we find that the DnPF can leverage the additional position information to substantially improve the prediction of the orientation component, for which no additional sensor information is available.
Notably, such sensor fusion cannot easily be achieved with end-to-end trained methods, which would, in general, require retraining to include the additional sensor information.

\subsection{Ablations}
\label{sec:ablation}

\begin{figure}[t]
\centering
\includegraphics[width=\linewidth]{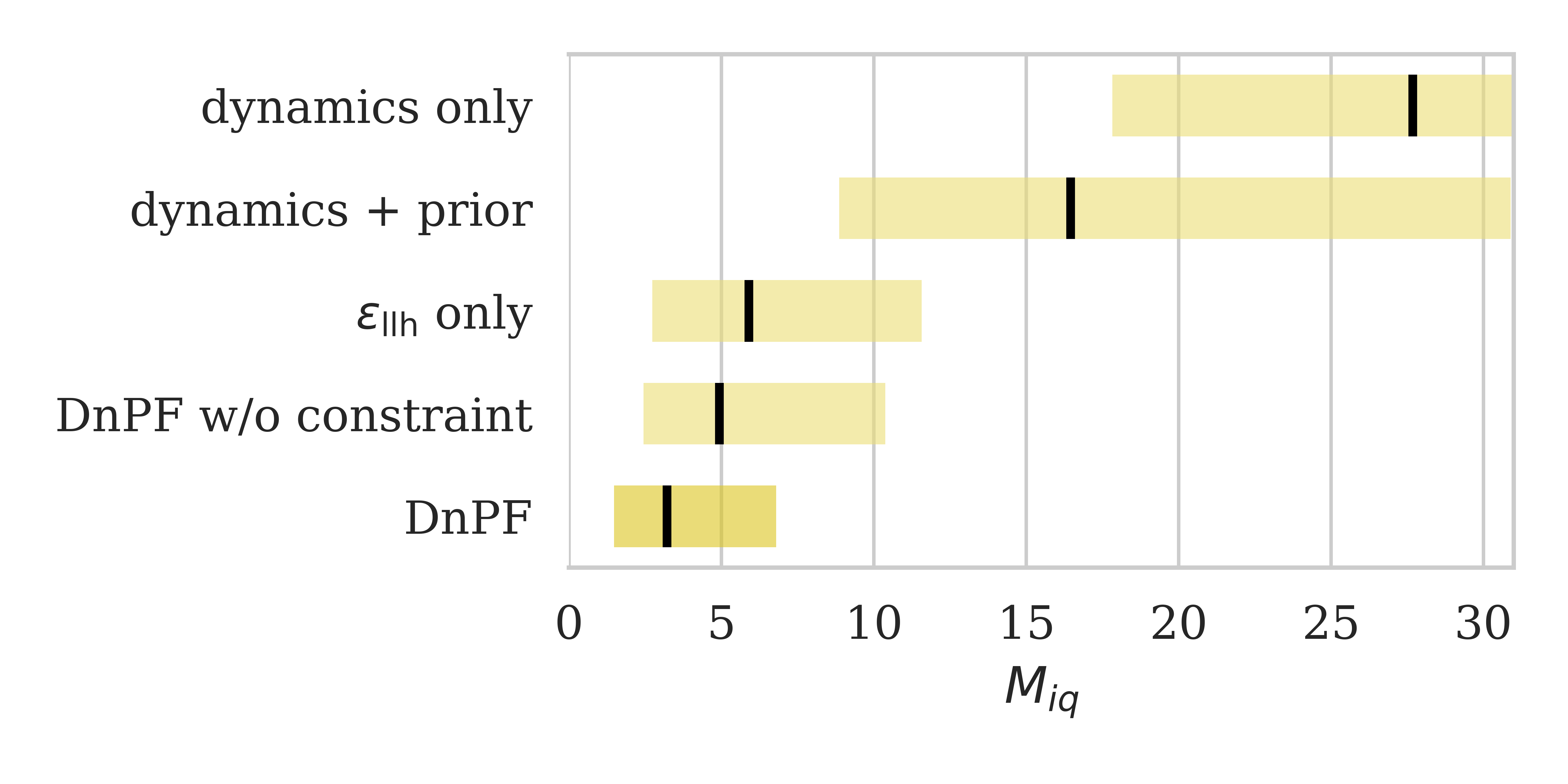}
\caption{Ablations on the DnPF inference procedure on the Multi-fingered Manipulation task. Shown are interquartile ranges (yellow) and medians (black) over 500 test sequences.
\textbf{dynamics only:} Unrolling the learned dynamics model $f$ for open-loop prediction leads to distributional drift and poor performance.
\textbf{dynamics + prior:} Denoising predictions from $f$ with the learned prior score $\nabla \log p_s(x_t)$ improves prediction robustness.
\textbf{$\epsilon_{\text{lh}}$ only:} Sampling particles using the learned likelihood score (no dynamics prior). 
\textbf{DnPF w/o constraint:} DnPF inference fusing dynamics and measurement likelihood but without the llh-constraint (from \cref{sec:constraint}).
\textbf{DnPF:} Full DnPF inference procedure.
}
\label{fig:ablation_plot}
\end{figure}

We perform a series of ablations to analyze the importance of different DnPF components on the Multi-fingered Manipulation task (\cref{fig:ablation_plot}).
First, recursively calling the learned dynamics model $f$ for open-loop prediction yields poor performance, likely due to distributional drift~\citep{Lutter2021-dv}.
Interestingly, combining learned dynamics models with the learned prior score $\nabla \log p_s(x_t)$ via warm-started few-step denoising ($\sim5$ steps here) mitigates this issue, facilitating more robust open-loop predictions.
Next, we consider drawing particles directly from the measurement likelihood $p(x_t | y_t)$ via diffusion sampling with learned $\epsilon_{\text{lh}}$, independently for each timestep.
Finally, for the full DnPF inference procedure combining learned dynamics and likelihood scores, the likelihood constraint introduced in \cref{sec:constraint} can significantly improve performance, as also shown qualitatively in \cref{fig:spin_plot}.

\subsection{Inference Runtime}

\begin{figure}[t]
\centering
\includegraphics[width=\linewidth]{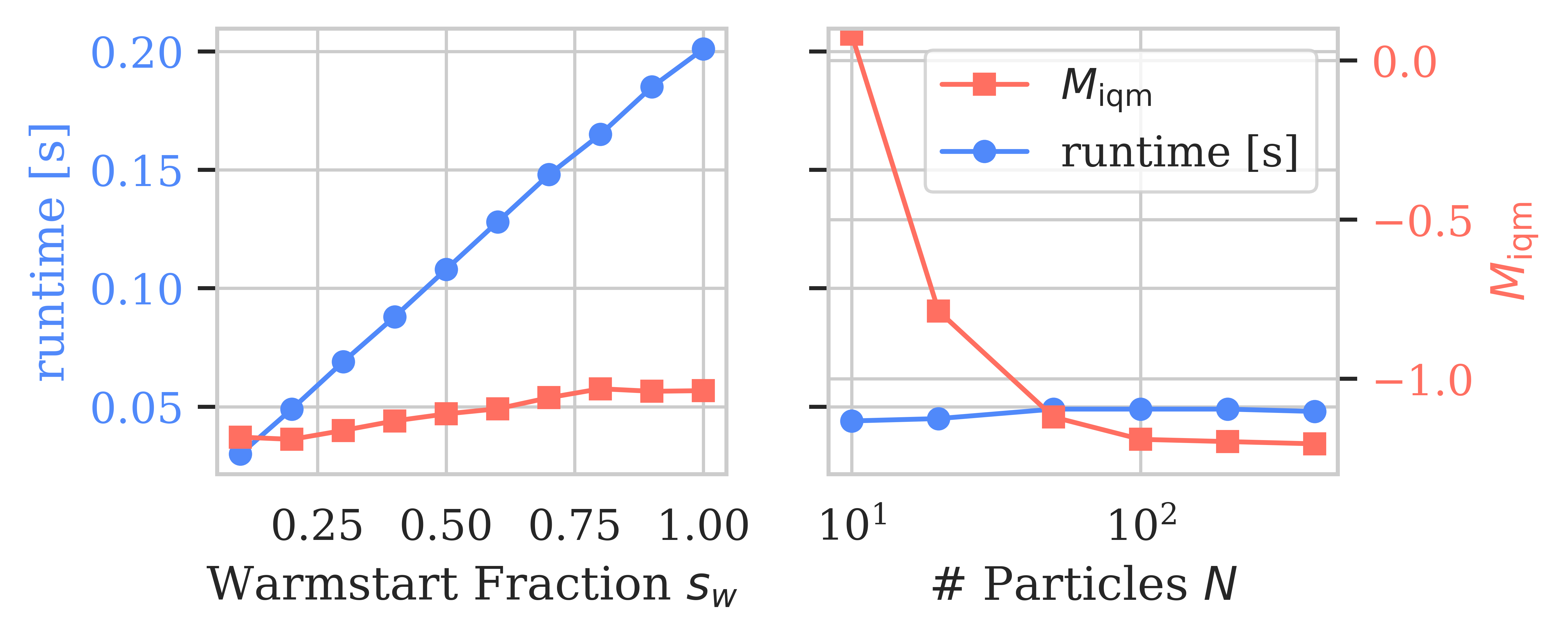}
\caption{Performance and inference runtime (GPU) analysis of DnPF with varying warm start fraction $s_{\mathrm{w}}$ (for fixed $N=100$) and varying number of particles $N$ (for fixed $s_{\mathrm{w}}=0.5$).}
\label{fig:runtime_plot}
\end{figure}

In \cref{fig:runtime_plot}, we analyze the inference runtime of DnPF per timestep on the \textit{Manipulator Spin} task with varying warm start fraction $s_{\mathrm{w}}$ and number of particles $N$.
As expected, performance generally improves with the number of particles. 
When inference is done on the GPU (Tesla T4 in this case), runtime is nearly independent of the number of particles up to the limit of parallel capacity.
For the warm start fraction $s_{\mathrm{w}}$, we observe a "sweet spot" with respect to prediction quality, which we find to be task-dependent, typically between $0.2$ and $0.5$. 
Runtime increases linearly with the number of denoising steps (and thus $s_{\mathrm{w}}$), as denoising steps are performed sequentially.
In the experiments, we use a maximum of $S=50$ denoising steps when $s_{\mathrm{w}}=1$.
In summary, DnPF with warm-starting allows for real-time inference on the studied tasks, even with a large number of particles.
\label{sec:runtime}

%===============================================================================

\section{Conclusions}
\label{sec:conclusion}
In this work, we proposed the Denoising Particle Filter (DnPF), a novel approach for state estimation in dynamical systems combining models learned via score matching with particle filtering.
On challenging state estimation tasks, DnPF performs competitively with end-to-end trained baselines, while being more robust to distributional shifts and allowing the integration of additional sensor models without retraining.
We believe the scalable training and modular design of DnPF make it a promising approach for many robotic applications. 
An interesting direction for future work is extending the framework to explicitly estimate unknown static parameters (e.g., friction coefficients), currently subsumed as noise in the learned models. 

\footnotesize
\bibliographystyle{IEEEtranN-modified}
\bibliography{IEEEabrv, references}

%%%%%%%%%%%%%%%%%%%%%%%%%%%%%%%%%%%%%%%%%%%%%%%%%%%%%%%%%%%%%%%%%%%%%%%%%%%%%%%%

\end{document}